\documentclass[preprint,twocolumn,10pt]{jmlr} 
\usepackage{xspace}

\newcommand{\tabspace}{\vspace{-0.5cm}}

\usepackage{amsmath,amsfonts,bm}

\usepackage{booktabs}
\usepackage{siunitx}
\usepackage{subcaption}
\usepackage{cleveref}

\usepackage[switch]{lineno}

\theorembodyfont{\upshape}
\theoremheaderfont{\scshape}
\theorempostheader{:}
\theoremsep{\newline}

 \title[wav2sleep]{wav2sleep: A Unified Multi-Modal Approach to Sleep Stage Classification from Physiological Signals}

 \author{%
  \Name{Jonathan F. Carter} \Email{jcarter@robots.ox.ac.uk}\\
  \Name{Lionel Tarassenko} \Email{lionel.tarassenko@eng.ox.ac.uk}\\
  \addr Institute of Biomedical Engineering, University of Oxford
 }

\begin{document}

\maketitle

\begin{abstract}
    Accurate classification of sleep stages from less obtrusive sensor measurements such as the electrocardiogram (ECG) or photoplethysmogram (PPG) could enable important applications in sleep medicine. Existing approaches to this problem have typically used deep learning models designed and trained to operate on one or more specific input signals. However, the datasets used to develop these models often do not contain the same sets of input signals. Some signals, particularly PPG, are much less prevalent than others, and this has previously been addressed with techniques such as transfer learning. Additionally, only training on one or more fixed modalities precludes cross-modal information transfer from other sources, which has proved valuable in other problem domains. To address this, we introduce wav2sleep, a unified model designed to operate on variable sets of input signals during training and inference. After jointly training on over 10,000 overnight recordings from six publicly available polysomnography datasets, including SHHS and MESA, wav2sleep outperforms existing sleep stage classification models across test-time input combinations including ECG, PPG, and respiratory signals.
\end{abstract}

\paragraph*{Data and Code Availability}
All datasets used in this work are available via the National Sleep Research Resource (NSRR, ~\citep{zhang_national_2018}). Training code, model weights, and processing pipelines for all datasets can be found here: \href{https://github.com/joncarter1/wav2sleep}{https://github.com/joncarter1/wav2sleep}.

\paragraph*{Institutional Review Board (IRB)}
The datasets used herein are publicly available via request to the NSRR and have been deidentified or fully anonymised, thus not requiring IRB approval. 

\section{Introduction}
\label{sec:wav2sleep:intro}
\begin{figure*}[htb]
\floatconts
{fig:wav2sleep:overview}
{\caption{\textbf{Overview of wav2sleep}. The model operates on \emph{sets} of time-series signals $\bm{X}_{1:T}$ to classify sleep stage sequences $y_{1:T}$. This enables it to be jointly trained on heterogeneous datasets, with different available signals, which are especially common in the healthcare domain. At inference time, the same model can be applied to any subset of the signals seen during training.}}
{\includegraphics[width=0.9\linewidth]{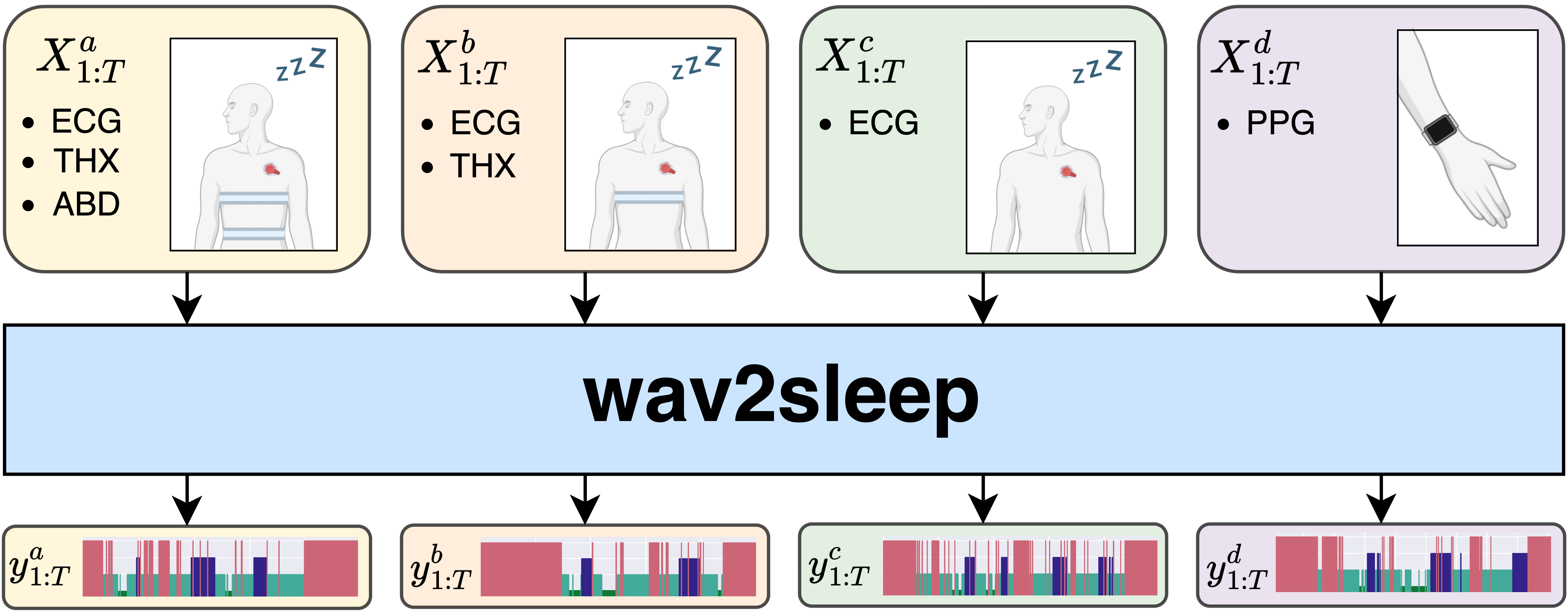}}
\end{figure*}
Quantitative analysis of sleep stages is important for applications such as the diagnosis of sleep disorders, the validation of sleep disorder medications, and the discovery of sleep biomarkers. This is typically done through a polysomnography (PSG) study following the American Academy of Sleep Medicine (AASM) guidelines~\citep{iber_aasm_2007}, during which a subject will undertake one or more nights of sleep whilst being continuously monitored using a variety of sensors.

After the study, a human expert will review the recording and assign each 30-second window of data (a sleep epoch) to one of 5 discrete sleep states defined by the AASM: Wake, N1 (light), N2 (intermediate), N3 (deep), or rapid eye movement (REM) sleep. This is traditionally done using signals such as the electroencephalogram (EEG), electrooculogram (EOG), and chin electromyogram (EMG), which are measured using electrodes attached to the subject's head.

The goal of sleep stage classification (sleep staging) algorithms is to automate this process using one or more input signals. Accurate sleep staging from inputs such as the photoplethysmogram (PPG), electrocardiogram (ECG), abdominal (ABD) and/or thoracic (THX) respiratory signals (see \Cref{fig:wav2sleep:overview}) is of particular interest since acquiring these signals is typically more comfortable for the subject and easier to set up than a full PSG study.

Sleep staging models have commonly been trained and evaluated using one or more fixed input modalities. However, the recorded signals often have high mutual information, meaning that information available from one input modality may be useful for learning to classify sleep stages from another.

To better leverage this information, we introduce wav2sleep, a unified model for sleep stage classification that operates on sets of physiological signals.  During training, the model can use all available input signals from each night of data, enabling it to be jointly trained across heterogeneous datasets, where the available signals may differ between recordings. This trait is particularly valuable for our problem domain of sleep staging since the set of recorded signals often varies, and many are often discarded due to poor signal quality.

After jointly training on multiple modalities, our model can use any subset of the original input signals during inference. We show that this unified approach outperforms existing methods across several test-time input modalities and datasets.

To summarise, our contributions are as follows:
\begin{itemize}
    \item We introduce wav2sleep, a novel deep learning architecture for multi-modal time-series that operates on sets of signals. This enables joint training on heterogeneous datasets, where the availability of signals can vary.
    \item We introduce a simple but effective stochastic masking procedure during training, which enables generalisation to an arbitrary subset of signals at test time.
    \item After training, we show that our single, unified model outperforms prior methods for sleep staging across a range of modalities. For example, outperforming existing ECG-based methods when using only the ECG at test time.
\end{itemize}

\section{Background and Motivation}\label{section:wav2sleep:motivation}
Prior work has shown that deep learning models can classify stages of sleep from brain signals recorded using the EEG with expert-level accuracy~\citep{phan_automatic_2022}. However, EEG signals are typically measured using electrodes attached to the patient’s scalp. To overcome this, other work has investigated classifying sleep stages using alternative, less obtrusive modalities such as the ECG~\citep{sridhar_deep_2020}, PPG~\citep{kotzen_sleepppg-net_2023}), and combinations of cardiac and respiratory signals~\citep{bakker_estimating_2021}.

\paragraph{Two-step learning} Automatic sleep staging from modalities such as the ECG or PPG is possible because these signals encode measures of physiological activity, such as heart rate variability (HRV), and these are known to be predictive of the sleep stage~\citep{shinar_automatic_2001}. Sleep staging from these signals can therefore be formulated as a two-step learning problem. First, we must find a mapping $f: \bm{x}\mapsto \bm{z}$ from the input signals $\bm{x}$ to relevant physiological features $\bm{z}$, then a mapping $g: \bm{z}\mapsto y$ from physiological features $\bm{z}$ to sleep stages $y$. For models trained end-to-end to classify sleep stages from input signals, these mappings $f$ and $g$ are jointly and implicitly learnt by a deep neural network. This has been shown to outperform methods where the physiological features $\bm{z}$ are derived using a manual feature engineering approach~\citep{kotzen_sleepppg-net_2023}, i.e. where the mapping $f$ is explicit and human-designed.

\paragraph{Transfer learning}
Sleep staging from modalities such as the PPG is of particular interest since they can be measured from ubiquitous wearables such as smartwatches~\citep{charlton_2023_2023}. However, PPG signals are less common in historical PSG datasets~\citep{radha_deep_2021}, hindering the use of deep learning methods which are notoriously data-hungry. To overcome this issue, transfer learning has been used to improve the performance of PPG-based models, by first pre-training using ECG data~\citep{radha_deep_2021, kotzen_sleepppg-net_2023}.

Transfer learning approaches from ECG to PPG signals work because these signals have high mutual information and morphological similarity. For example, both signals encode HRV and respiratory rate variability (RRV) information. This fact is explicitly used by \cite{radha_deep_2021}, who first pre-train a model using HRV features derived from the ECG, then fine-tune using HRV features derived from PPG data. Using the two-step learning formulation introduced at the start of this section, this approach can be thought of as explicitly mapping each input signal $\bm{x}_i$ to a modality-agnostic feature space $\bm{z}$ via a modality-dependent mapping $f_i$.

\paragraph{Modality-agnostic learning} Ideally, modality-agnostic relationships such as the link between HRV and deep sleep~\citep{shinar_automatic_2001}, or between RRV and REM sleep~\citep{kantelhardt_breathing_2003}, should not need to be learnt from a specific modality, a fact which is already partially exploited by transfer learning approaches. However, transfer learning is prone to the problem of catastrophic forgetting~\citep{kemker_measuring_2018} i.e. where after training on new data, information learnt from the old dataset is `forgotten'. Hence, even if our end goal is to produce a specialised model for a particular modality, we will show that it is beneficial to use a model that is jointly trained across multiple modalities, rather than directly training or using a transfer learning approach.

\paragraph{Multi-modal learning}
Multi-modal sleep staging methods have varied in their approaches to combining cross-modal information. For example, the model proposed by \cite{chambon_deep_2018} independently turns each input signal into features before passing them to a classifier (late fusion). In contrast, models such as SeqSleepNet~\citep{phan_seqsleepnet_2019} concatenate modalities at the input level (early fusion). In the middle ground, separately extracting features from each modality for each 30-second sleep epoch, then fusing and jointly modelling sequential information, is also a common approach, e.g. via concatenation~\citep{bakker_estimating_2021, carter_sleepvst_2024} or cross-modal attention~\citep{wang_caresleepnet_2024, pradeepkumar_towards_2024}.

For EEG and EOG signals, \cite{kontras_core-sleep_2024} handle heterogeneity by simultaneously training both uni-modal and multi-modal models using an alignment loss, with the uni-modal model used when only one signal is available. This is shown to improve uni-modal performance at test time. Alternatively, MaskSleepNet~\citep{zhu_masksleepnet_2023}, designed for EEG, EOG and/or EMG signals, handles heterogeneity by zero-padding the EOG and/or EMG input signals, which is done randomly during training.

\paragraph{Objective} Our aim is to learn a modality-agnostic intermediate representation of the input signals because we hypothesise this can enable more robust representation learning, improving sleep staging performance. However, we want to maintain the ability to perform end-to-end deep learning from the raw input signals, since this has led to improved performance not just in sleep staging, but across numerous problem domains~\citep{goodfellow_deep_2016}. Finally, we desire a model that can be jointly trained across heterogeneous datasets, improving the diversity of input data, and avoiding the problem of catastrophic forgetting. In Sections \ref{section:wav2sleep:arch} and \ref{section:wav2sleep:setup}, we describe the wav2sleep model which addresses these challenges, and discuss its advantages over prior methods. In \Cref{section:wav2sleep:results}, we empirically validate its effectiveness in sleep staging across a range of test-time modalities.

\section{Model Architecture}\label{section:wav2sleep:arch}
\begin{figure*}[htbp]
\floatconts
{fig:wav2sleep:model}
{\caption{\textbf{wav2sleep architecture for sets of signals.} (\emph{a}) Each input signal $x^{i}_{1:kT}$ from modality $i\in\mathcal{S}$ is passed to a CNN to form a sequence of feature vectors $\bm{z}^i_{1:T}$. (\emph{b}) For each time-step $t$, a transformer encoder turns the set of features into a single aggregate feature vector $\bm{z}_t$ using a CLS token~\citep{devlin_bert_2019}. (\emph{c}) A dilated CNN mixes sequential information to classify sleep stage output sequences $y_{1:T}$}
}
{
\subfigure[Signal Encoders][b]{\label{fig:wav2sleep:sig_enc}%
  \includegraphics[width=0.3\linewidth]{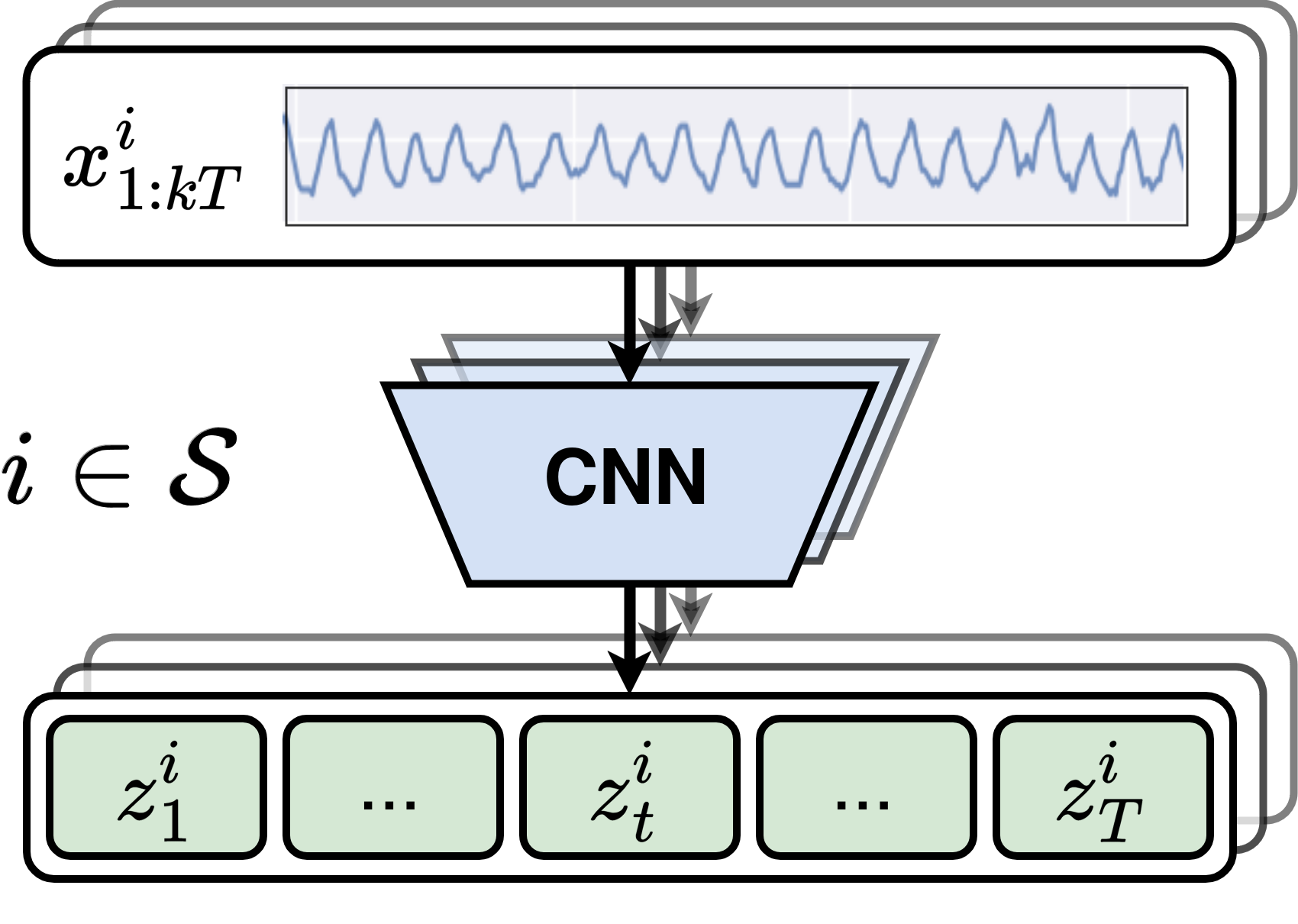}}%
\hfill
\subfigure[Epoch Mixer][b]{\label{fig:wav2sleep:epoch_mixer}%
  \includegraphics[width=0.29\linewidth]{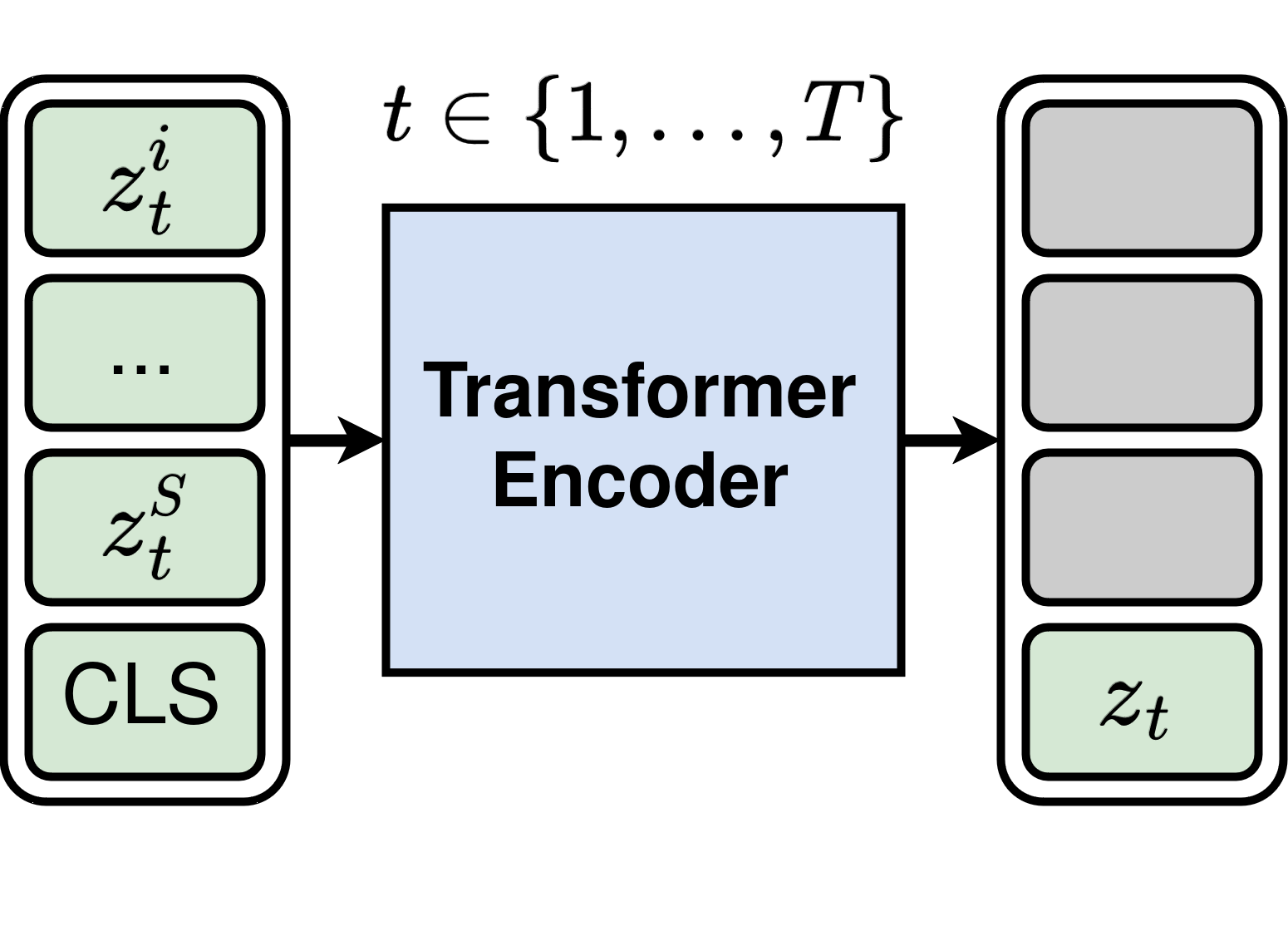}}
\hfill
\subfigure[Sequence Mixer][b]{\label{fig:wav2sleep:seq_mixer}%
  \includegraphics[width=0.3\linewidth]{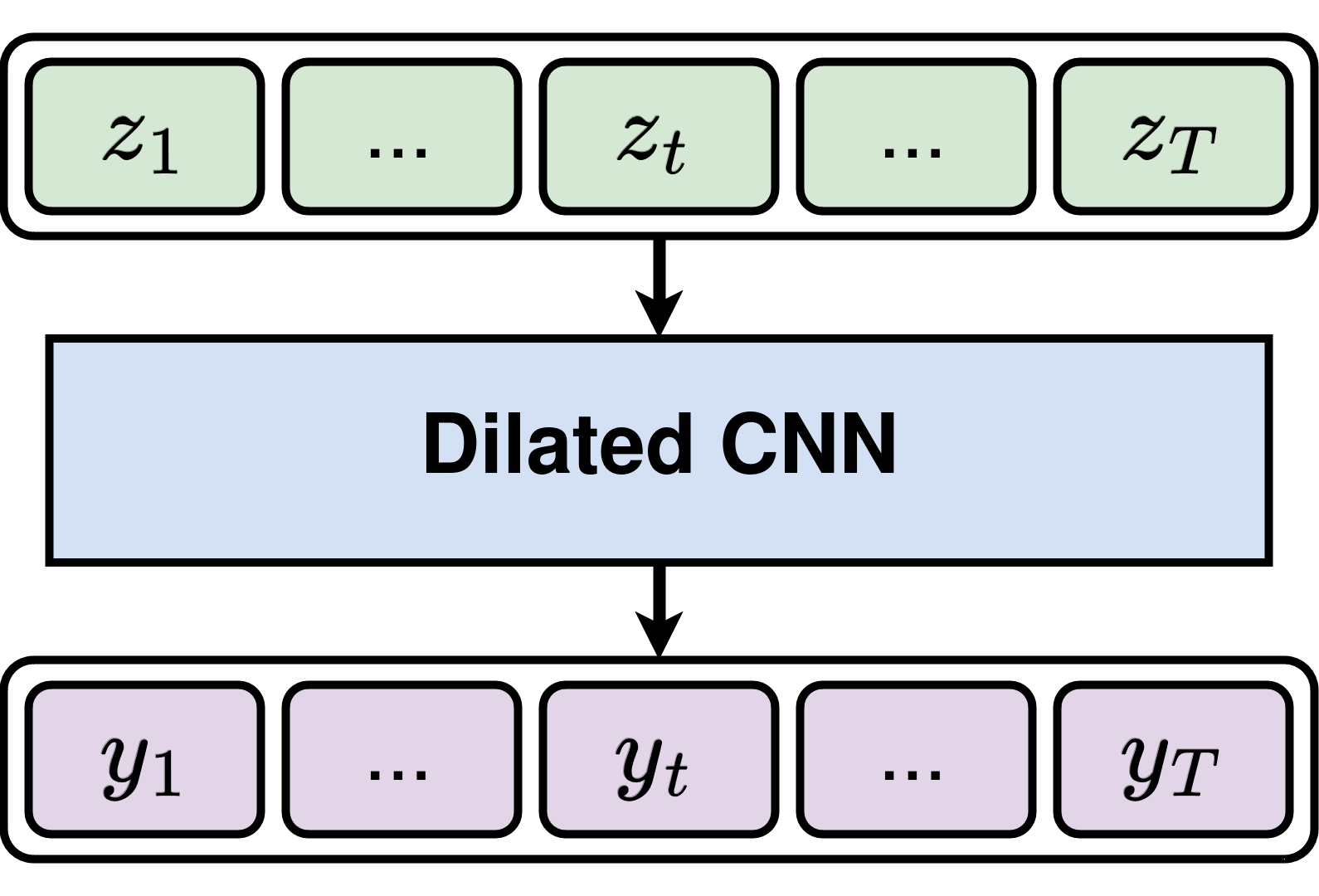}}
}
\end{figure*}
In this section, we describe the architecture of the wav2sleep model, which turns sets of time-series signals spanning multiple hours into sleep stage classifications for each 30-second sleep epoch. The model architecture,  illustrated in \Cref{fig:wav2sleep:model}, consists of three high-level components:
\begin{enumerate}
\item  \textbf{Signal Encoders}, which independently extract features for each signal in the input set $\bm{X}_{1:T}$.
\item \textbf{Epoch Mixer}, which fuses cross-modal information into a unified representation $\bm{z}_t$ for each sleep epoch.
\item \textbf{Sequence Mixer}, which mixes information temporally to classify sleep stages $y_{1:T}$.
\end{enumerate}

\subsection{Signal Encoders}
The model first turns the set of continuous 1D input signals $\bm{X}_{1:T} = \{x^i_{1:kT} |\,i \in \mathcal{S} \}$ into a set of feature vector sequences $\bm{Z}_{1:T} = \{\bm{z}^i_{1:T} |\,i \in \mathcal{S}\}$, where $\bm{z}^i_t$ denotes the feature vector for modality $i$ for sleep epoch $t$, $k$ denotes the relative sampling rate of each signal, and $\mathcal{S}$ denotes the set of available modalities e.g. ECG and PPG signals. We use separate CNN encoders for each input modality, which follow the design of the early layers of SleepPPG-Net~\citep{kotzen_sleepppg-net_2023}. These consist of a stack of residual layers~\citep{he_deep_2016}, each containing three convolutional layers followed by a max pooling layer to downsample the signal by a factor of 2. The residual layers are followed by a reshape operation and a time-distributed dense layer to produce the sequence of feature vectors $\bm{z}^i_{1:T}$.

\subsection{Epoch Mixer}
Having independently transformed each modality $i$ into a sequence of feature vectors $\bm{z}^i_{1:T}$, we next fuse information from the set of modalities to provide a single unified representation $\bm{z}_t$ for each sleep epoch i.e. to complete the mapping $f$ described in \Cref{section:wav2sleep:motivation}. We use a transformer encoder~\citep{vaswani_attention_2017} to do this, providing the transformer with an extra learnable vector, i.e. a CLS token~\citep{devlin_bert_2019, dosovitskiy_image_2020}, and using the output at that position as our unified feature vector. This design straightforwardly handles a varying number of input modalities during training and inference whilst keeping the dimensionality of the fused feature sequence $\bm{z}_{1:T}$ fixed.

\subsection{Sequence Mixer}
The feature vectors $\bm{z}_{1:T}$ are passed to the sequence mixer, which mixes sequential information to produce sleep stage outputs $y_{1:T}$. This is a desirable property since sleep exhibits long-range time-series structures such as sleep cycles~\citep{patel_physiology_2022}. We use a dilated CNN design as previously used by \cite{sridhar_deep_2020, kotzen_sleepppg-net_2023}. This consists of multiple blocks of dilated convolutional layers where the dilation doubles at each layer, meaning that the size of the model's receptive field increases exponentially with network depth.

\subsection{Advantages}
Returning to the two-step learning formulation ($f$ and $g$) introduced in \Cref{section:wav2sleep:motivation}, the wav2sleep architecture has two key advantages:
\begin{enumerate}
    \item Because it operates on \emph{sets} of input signals, the model can be trained on heterogeneous datasets, increasing the variety of data available in terms of both the input modalities (for learning $f$) \emph{and} physiology (for learning $g$).
    \item By training on all available modalities \emph{jointly}, this should lead to more robust learning in the presence of noise, by avoiding shortcut learning when one or more modalities are corrupted.
\end{enumerate}
Using a unified model also has practical advantages, since only a single model needs to be trained, validated and deployed, reducing operational complexity for real-world applications.

\section{Experimental Set-up}\label{section:wav2sleep:setup}
\subsection{Datasets and Preprocessing}\label{section:wav2sleep:setup:preprocessing}
We use 7 PSG datasets available from the National Sleep Research Resource~\citep{zhang_national_2018}: SHHS~\citep{quan_sleep_1997}, MESA~\citep{chen_racialethnic_2015}, CFS~\citep{redline_familial_1995}, MROS~\citep{blackwell_associations_2011}, CHAT~\citep{marcus_randomized_2013}, CCSHS~\citep{rosen_prevalence_2003}, and WSC~\citep{young_burden_2009}. Demographic information for the datasets used is provided in \Cref{table:wav2sleep:demographics}. Collectively, these datasets contain over 15,000 pairs of overnight polysomnography recordings and expert-annotated sleep stages. Notably, there is significant variation in patient demographics. For example, the SHHS, MESA and WSC datasets are mostly comprised of recordings from older adults with high apnea-hypopnea indices (sleep-disordered breathing). In contrast, the CCSHS and CHAT datasets both contain PSG recordings from children. Joint training across all datasets exposes the model to a wider variety of contact sensors (makes, models etc.) and individual physiological variations.

\begin{table*}[htb]
\floatconts
{table:wav2sleep:demographics}
{\caption{Demographics, dataset split sizes, and signal availability for the PSG datasets used.}}
{\tabspace\begin{tabular}{@{}llllllll@{}}
\toprule
Characteristic & SHHS & MESA & WSC & CHAT & CFS & CCSHS & MROS\\ \midrule
\textbf{Demographics}$^\dag$\\
Age, mean & 65.2 & 69.6 & 59.8 & 7.2 & 41.4 & 17.7 & 78.7\\
Sex, m:f & 0.88:1 & 0.87:1 & 1.17:1 & 0.94:1 & 0.81:1 & 1.02:1 & 1:0\\
AHI$^\ddag$, mean & 15.2 & 20.4 & 20.0 & 5.5 & 13.2 & 1.5 & 18.3\\
\textbf{Splits},\,\,\small{N\,(\%)}\\
Train & 6441\,\small{(81\%)} & 1541\,\small{(84\%)} & 1380\,\small{(65\%)} & 1132\,\small{(79\%)} & 452\,\small{(75\%)} & 272\,\small{(64\%)} & 0\\
Validation & 500\,\,\,\,\small{(6\%)} & 100\,\,\,\,\small{(5\%)} & 250\,\,\,\,\small{(12\%)} & 100\,\,\,\,\small{(7\%)} & 50\,\,\,\,\small{(8\%)} & 50\,\,\,\,\small{(12\%)} & 0\\
Test & 1000\,\small{(13\%)} & 200\,\,\,\,\small{(11\%)} & 500\,\,\,\,\small{(23\%)} & 200\,\,\,\,\small{(14\%)} & 100\,\small{(17\%)} & 100\,\small{(24\%)} & 1000\\
\textbf{Signals}\\
\small{ECG/ABD/THX} & 7941 & 1841 & 2130 & 1432 & 602 & 422 & 1000\\
\small{PPG} & 0 & 1841 & 0 & 1139 & 284 & 422 & 0\\
\bottomrule
\multicolumn{7}{l}{\scriptsize{$^\dag$Calculated from NSRR harmonized variables (nsrr\_age, nsrr\_sex, nsrr\_ahi\_hp3u). $^\ddag$AHI - apnea-Hypopnoea Index.}}
\end{tabular}
}
\end{table*}
There is also variation in the availability of signals between the datasets. For example, recordings from MESA and CCSHS, and some from the CFS and CHAT datasets, contain a PPG signal, but recordings from the other datasets do not.  Where available, we used the ABD and THX respiratory signals, the PPG signal, and the ECG signal from each recording.

\paragraph{Dataset splits}
Although numerous prior works have explored the problem of sleep staging on the datasets used, there are no widely-established fixed training, validation and test partitions. We therefore establish new splits for all data sets, excluding nights that have not been annotated with multiple sleep stages as done in prior work~\citep{phan_xsleepnet_2022}. For datasets that contain multiple recordings from a single participant, we ensured that no participant appeared in both the test set and either the training or validation sets. No other exclusion criteria--such as signal quality heuristics~\citep{jones_expert-level_2024}--were explicitly used since one of the key aims of training on multiple modalities jointly is to improve robustness to noise on any particular channel.

The size of our training, validation and test set splits for each dataset are listed in \Cref{table:wav2sleep:demographics}. These were chosen to be in line with those used in prior work, e.g.~\citep{sridhar_deep_2020}. Our splits were carefully constructed to additionally allow evaluation on the aggregated test set proposed by \cite{jones_expert-level_2024}, which uses multiple PSG datasets to create a test set that approximately matches the 2022 US census demographics. Throughout the remainder of this paper, we refer to this as the `Census' test set. More detail on the construction of our training, validation and test sets is provided in \Cref{section:wav2sleep:appendix:dataset_processing}.

\paragraph{Preprocessing}
We minimally processed all signals using a similar process to that described by \cite{kotzen_sleepppg-net_2023}, padding or truncating each recording to 10 h (i.e. sequence length $T=1200$), re-sampling each signal to the same frequency across recordings, and applying unit normalisation. The ECG and PPG signals were resampled such that each 30-second sleep epoch consisted of $k=1024$ data points ($\approx 34$ Hz), which simplifies temporal alignment during pooling operations within the convolutional layers of the signal encoders. Since respiratory signals are generally sampled at a lower frequency during PSG recordings (e.g. 5-10 Hz in SHHS), the ABD and THX signals were resampled to a lower frequency of $k=256$ data points per sleep epoch ($\approx 8$ Hz), reducing the computational and memory requirements of the model during training and inference.

\subsection{Model training}\label{section:wav2sleep:training}
All models were trained to minimise the cross-entropy loss between expert-annotated sleep stages and model outputs using the AdamW optimiser~\citep{loshchilov_decoupled_2019} with a batch size of 16 and weight decay of $10^{-2}$. For the learning rate schedule, we used a linear warm-up of 2000 steps to a maximum learning rate $\epsilon=10^{-3}$ followed by an exponential decay to zero. Training continued until there was no decrease in the loss on the validation set for 5 epochs, which typically required around 30 epochs in total. Further training details can be found in \Cref{section:wav2sleep:appendix:training}. The checkpoint which resulted in the lowest validation loss was restored for evaluation. Model hyper-parameters were tuned using the validation sets before evaluation on the test sets took place.

\paragraph{Augmentation}
As noted by \cite{jones_expert-level_2024}, signals such as the ECG are sometimes inverted due to electrodes being connected the wrong way around. To improve robustness, all signals were randomly inverted (multiplied by -1) with a 50\% probability during training.

\subsection{Model hyper-parameters}
Hyper-parameters for the wav2sleep model are listed in \Cref{table:wav2sleep:hyperparams}. In each signal encoder, the number of residual layers, and the number of channels in each layer, were chosen so that the resulting feature dimension is independent of the relative sampling rate $k$. For simplicity, we retained this feature dimension ($\text{dim}(\bm{z}^i_t)=\text{dim}(\bm{z}_t)=128$) throughout the remainder of the model. Additional architecture details can be found in \Cref{section:appendix:wav2sleep:model_design}.
\begin{table}[htb]
\floatconts
{table:wav2sleep:hyperparams}
{\caption{wav2sleep model hyper-parameters.}}
{\tabspace\small
\begin{tabular}{@{}ll@{}}
\toprule
Hyper-parameter & Value\\ \midrule
\textbf{Global} & \\
Feature dimension & 128\\
Activation function & GELU$^{\text{\dag}}$\\
Dropout & 0.1\\
\textbf{Signal Encoders} & \\
Kernel size & 3\\
Channels ($k=256$) & \footnotesize{(16,32,64,64,128,128)}\\
Channels ($k=1024$) & \footnotesize{(16,16,32,32,64,64,128,128)}\\
\textbf{Epoch Mixer} & \\
Transformer layers & 2\\
Hidden dimension & 512\\
Attention heads & 8\\
\textbf{Sequence Mixer} & \\
Dilated blocks & 2\\
Kernel size & 7\\
Dilations (per block) & \small{(1,2,4,8,16,32)}\\
\bottomrule
\multicolumn{2}{@{}l@{}}{\scriptsize$^{\text{\dag}}$\cite{hendrycks_gaussian_2023}.} 
\end{tabular}
}
\end{table}

\subsection{Stochastic masking}
During training, to handle differences in the available modalities within a batch, we padded unavailable signals and added a mask to the attention matrices of the epoch mixer. To aid test-time generalisation to a subset of modalities, we randomly sampled a subset of the available signals for each recording via additional masking of the attention matrix. Where available, the input signals were masked with the following probabilities:
\begin{align*}
p(m_\text{ABD})=0.7\quad & p(m_\text{THX})=0.7\\
p(m_\text{ECG})=0.5\quad & p(m_\text{PPG})=0.1
\end{align*}
These values were intuitively chosen so that the higher frequency ECG and PPG signals were less likely to be masked, and to increase the prevalence of the scarcer PPG signal.

\begin{figure}[htbp]
\floatconts{fig:masking}
{\caption{\textbf{Stochastic masking.} During training, we sample a random subset of the available modalities for each night of data. To retain a fixed batch shape, we pad unavailable modalities and apply a mask to the self-attention matrices of the epoch mixer.}}
{\includegraphics[width=0.97\linewidth]{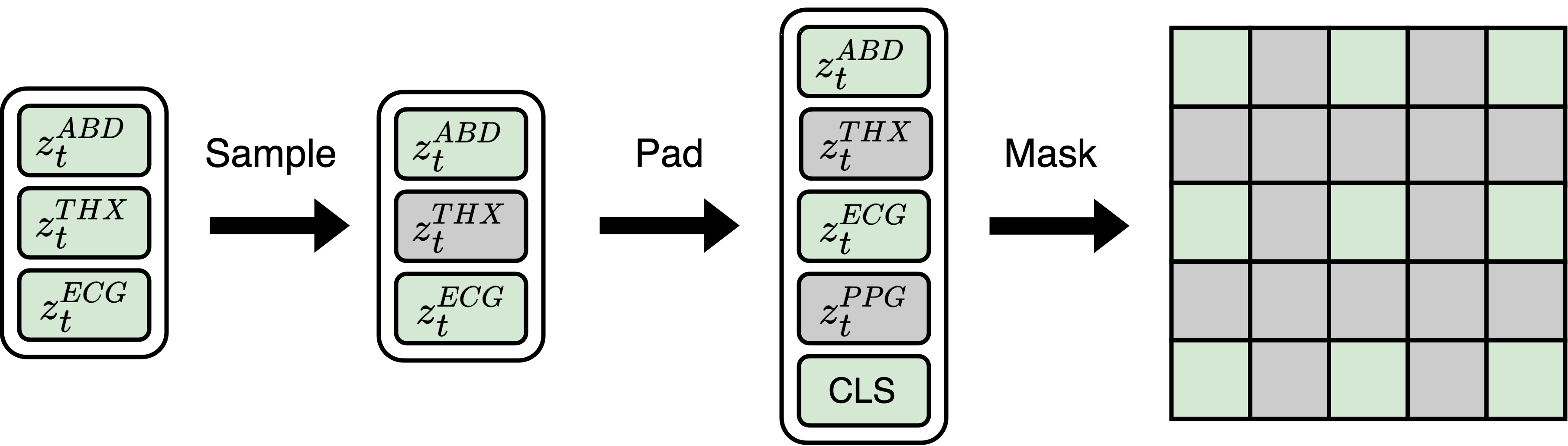}}
\end{figure}

Our stochastic masking process draws inspiration from masked modeling approaches e.g.~\citep{baevski_wav2vec_2020, he_masked_2021}, and is similar to hierarchical channel sampling (HCS,~\citep{bao_channel_2024}) from the visual domain. However, the practical implementation of HCS requires all samples within a batch to have the same available channels, to retain a fixed batch shape during training. Our approach (illustrated in \Cref{fig:masking}) allows heterogeneity \emph{within} batches, which simplifies training on heterogeneous datasets. During inference, we simply pass only the available signals to the transformer i.e. there is no need for any padding or manipulation of the attention mechanism to handle different numbers of input signals at test time.
\section{Results and Discussion}
\label{section:wav2sleep:results}
In this section, we report the performance of our model in four-class sleep staging, merging N1 and N2 into a single `Light' sleep class as commonly done in prior work. We report total Cohen's $\kappa$ ($\kappa_{T}$) and accuracy ($\text{Ac}_{T}$) calculated over all sleep epochs in each test set. All results are averages over three training runs using different random seeds. Our full set of results for all dataset-modality combinations evaluated can be found in \Cref{section:wav2sleep:appendix:results:modalities}.

\subsection{Cross-modal learning}
In \Cref{table:wav2sleep:ppg}, we compare the performance of three approaches to PPG-based sleep staging:
\begin{enumerate}
    \item Direct training on (scarce) PPG signals.
    \item Transfer learning from ECG to PPG signals.
    \item Joint training on all available modalities.
\end{enumerate} We additionally compare with a re-implementation of SleepPPG-Net~\citep{kotzen_sleepppg-net_2023}, trained using the same splits and learning procedure as our model. Using transfer learning ($\mathcal{S}_{\,\text{Train}}=\text{ECG}\rightarrow\text{PPG}$), we pre-train using the ECG signal, then fine-tune using the PPG signal, resuming the learning rate schedule. Across datasets, we find that our joint training approach with stochastic masking consistently leads to better performance than either direct training or transfer learning for the scarce PPG modality.
\setlength{\tabcolsep}{0.2em}
\begin{table}[htb]
\floatconts
{table:wav2sleep:ppg}
{\caption{Performance ($\kappa_{T}$) for $\mathcal{S}_{\,\text{Test}}=\text{PPG}$.}}
{\tabspace\footnotesize
    \begin{tabular}{@{}llcccc@{}}
    \toprule
    & & \multicolumn{4}{c}{Dataset}\\
    Model & $\mathcal{S}_{\,\text{Train}}$ & MESA & CHAT & CFS & CCSHS\\ \midrule
    \footnotesize{SleepPPG-Net} & \footnotesize{PPG} & 0.713 & 0.757 & 0.731 & 0.803\\
    & \footnotesize{ECG$\rightarrow$PPG} & 0.724 & 0.767 & 0.746 & 0.808  \\ 
    wav2sleep  & \footnotesize{PPG} & 0.728 & 0.777 & 0.751 & 0.817\\
      & \footnotesize{ECG$\rightarrow$PPG} & 0.732 & 0.779 & 0.754 & 0.811\\
      & All$^\dag$ & \textbf{0.742} & \textbf{0.793} & \textbf{0.763} & \textbf{0.832}\\
    \bottomrule
    \multicolumn{6}{@{}l@{}}{\scriptsize $^\dag$ABD+THX+ECG (+PPG for MESA, CHAT, CFS, CCSHS)}
    \end{tabular}
    }
\end{table}

Similarly, \Cref{table:wav2sleep:ecg} compares the performance of direct and joint training for sleep staging using the (abundant) ECG signal. For the SHHS and WSC datasets, and for the completely held-out MROS dataset, joint training resulted in the best performance. However, for some datasets, we found that joint training without the PPG signal ($\mathcal{S}_{\,\text{Train}}=\text{\small{No PPG}}$) resulted in better ECG-only performance. This indicates that cross-modal learning from respiratory signals to ECG was able to occur, but that there is a trade-off between learning from ECG and PPG signals for some datasets. This is a limitation of our work which is further discussed in \Cref{section:wav2sleep:appendix:results:tradeoff}.

\begin{table}[htbp]
\floatconts
{table:wav2sleep:ecg}
{\caption{Performance ($\kappa_{T}$) for $\mathcal{S}_{\,\text{Test}}=\text{ECG}$.}}
{\tabspace
    \footnotesize
    \begin{tabular}{@{}llccccc@{}}
    \toprule
    & &  \multicolumn{5}{c}{Dataset}\\
    Model & $\mathcal{S}_{\,\text{Train}}$ & SHHS & WSC & CFS & Census & MROS\\ \midrule
    \scriptsize{SleepPPG-Net} & \footnotesize{ECG} & 0.722 & 0.671 & 0.762 & 0.765 & 0.712\\
    wav2sleep & \footnotesize{ECG} & 0.733 & 0.683 & 0.785 & 0.786 & 0.746\\
    & \scriptsize{No\,PPG} & 0.738 & 0.686 & \textbf{0.789} & \textbf{0.792} & 0.748\\
    & \footnotesize{All} & \textbf{0.739} & \textbf{0.689} & 0.784 & 0.783 & \textbf{0.750}\\
    \bottomrule
    \multicolumn{7}{@{}l@{}}{\scriptsize $^\dag$ABD+THX+ECG (+PPG for MESA, CHAT, CFS, CCSHS)}
    \end{tabular}
}
\end{table}

\subsection{Varying modalities}
\Cref{fig:wav2sleep:supp:cmats} shows confusion matrices between expert-annotated sleep stages and model outputs for different test-time modalities using the Census test set, aggregated over all sleep epochs. Here we see that the addition of breathing signals (ABD, THX) is particularly helpful in distinguishing both Wake and REM from Light (N1+N2) sleep. Using just the ECG and THX signals, we obtain a Cohen's $\kappa$ of 0.812. Whilst caution should be taken when interpreting $\kappa$ values, notably, using the rule-of-thumb proposed by~\cite{landis_measurement_1977} this corresponds to `almost perfect' agreement with the expert-annotated sleep stages.

In \Cref{fig:wav2sleep:extra_modalities}, we plot the performance of the wav2sleep model for different age ranges and test-time modalities $\mathcal{S}_{\text{Test}}$. We observe good performance across age ranges, and that using more modalities consistently leads to improved performance, particularly by reducing the quantity and severity of outliers.

\begin{figure*}[pt]
\floatconts
  {fig:wav2sleep:supp:cmats}
  {\caption{Sleep stage confusion matrices for varying $\mathcal{S}_{\text{Test}}$ on the Census test set.}}
  {%
    \subfigure[$\mathcal{S}_{\text{Test}}=\text{ECG}$]{\label{fig:wav2sleep:supp:ecg}%
      \includegraphics[width=0.33\linewidth]{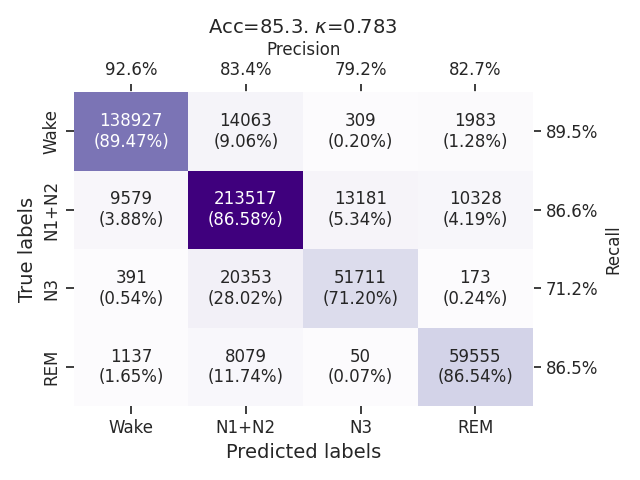}}%
    \hfill
    \subfigure[$\mathcal{S}_{\text{Test}}=\{\text{ECG, THX}\}$]{\label{fig:wav2sleep:supp:ecg_thx}%
      \includegraphics[width=0.33\linewidth]{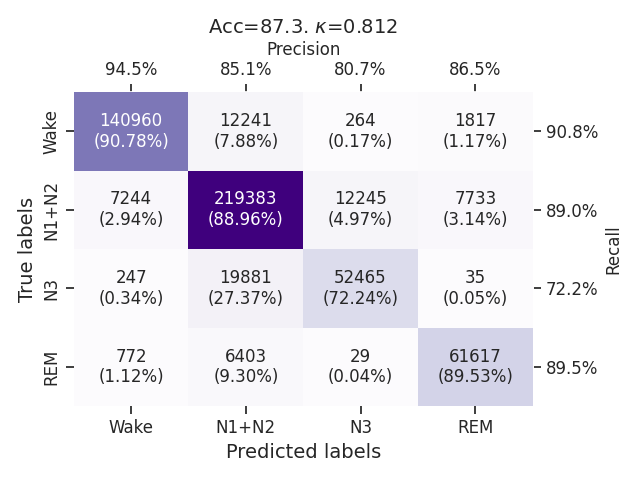}}
    \hfill
    \subfigure[$\mathcal{S}_{\text{Test}}=\text{All}$]{\label{fig:wav2sleep:supp:all}%
      \includegraphics[width=0.33\linewidth]{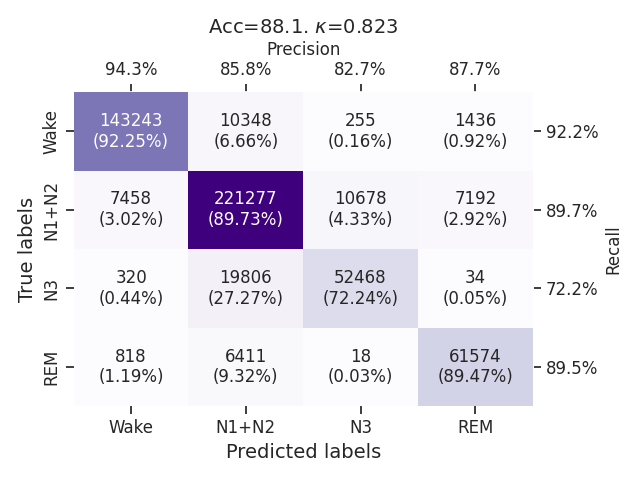}}
  }
\end{figure*}

These outliers are often caused by noise on a particular signal (see \Cref{section:wav2sleep:appendix:results:noise}), but can also be caused by specific physiological conditions. Notably, we found that when using the ECG as the sole input, performance improved with apnea severity for subjects with cardiac arrhythmia (see \Cref{section:wav2sleep:appendix:hypnograms}). This is in contrast to the general trend seen in prior work that the performance of sleep staging models tends to decrease with apnea severity~\citep{korkalainen_accurate_2020}.

\begin{figure}[hp]
\floatconts
  {fig:wav2sleep:extra_modalities}
  {\caption{Performance ($\kappa_{T}$) of wav2sleep against age for varying $\mathcal{S}_{\text{Test}}$ on the Census dataset.}}
{\includegraphics[width=1.0\columnwidth]{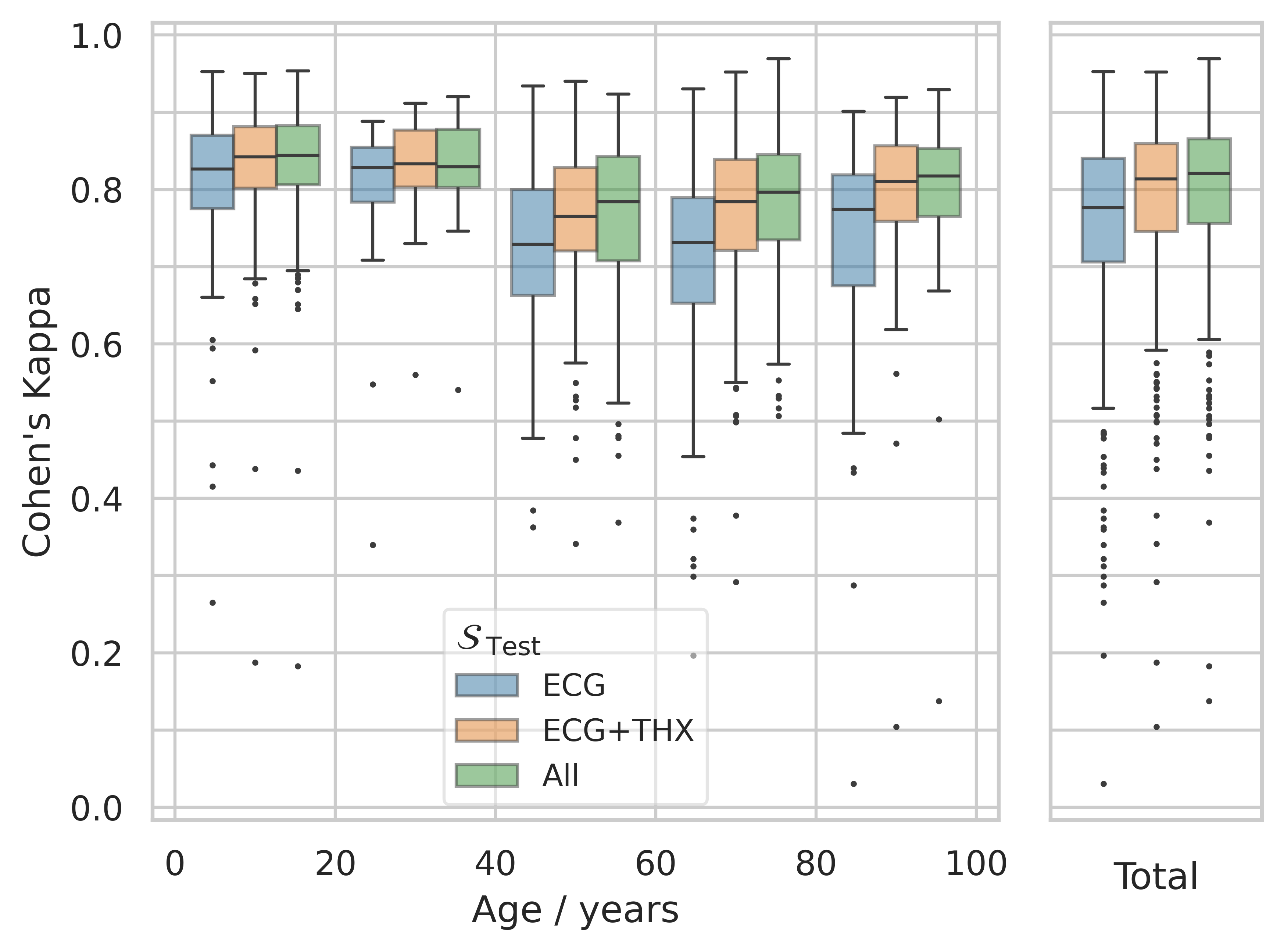}}
\end{figure}

This highlights how, during real-world deployment, the set of input modalities (contact sensors) may need to be chosen in a patient-specific manner to ensure an expected level of accuracy in the presence of physiological confounders. The use of a single, unified model such as wav2sleep can help to simplify such a process.

\subsection{Stochastic masking}
\Cref{fig:wav2sleep:masking_radar} shows the performance of wav2sleep for various dataset-modality combinations with and without the use of stochastic masking during training. Here we can see that stochastic masking is essential for generalisation to subsets of modalities at test-time, whilst maintaining equivalent performance when using all modalities.
\begin{figure}[htbp]
\floatconts
  {fig:wav2sleep:masking_radar}
  {\caption{Performance ($\kappa_{T}$) of wav2sleep for various dataset-modality combinations with and without stochastic masking (SM) during training.}}
{\includegraphics[width=1.0\columnwidth]{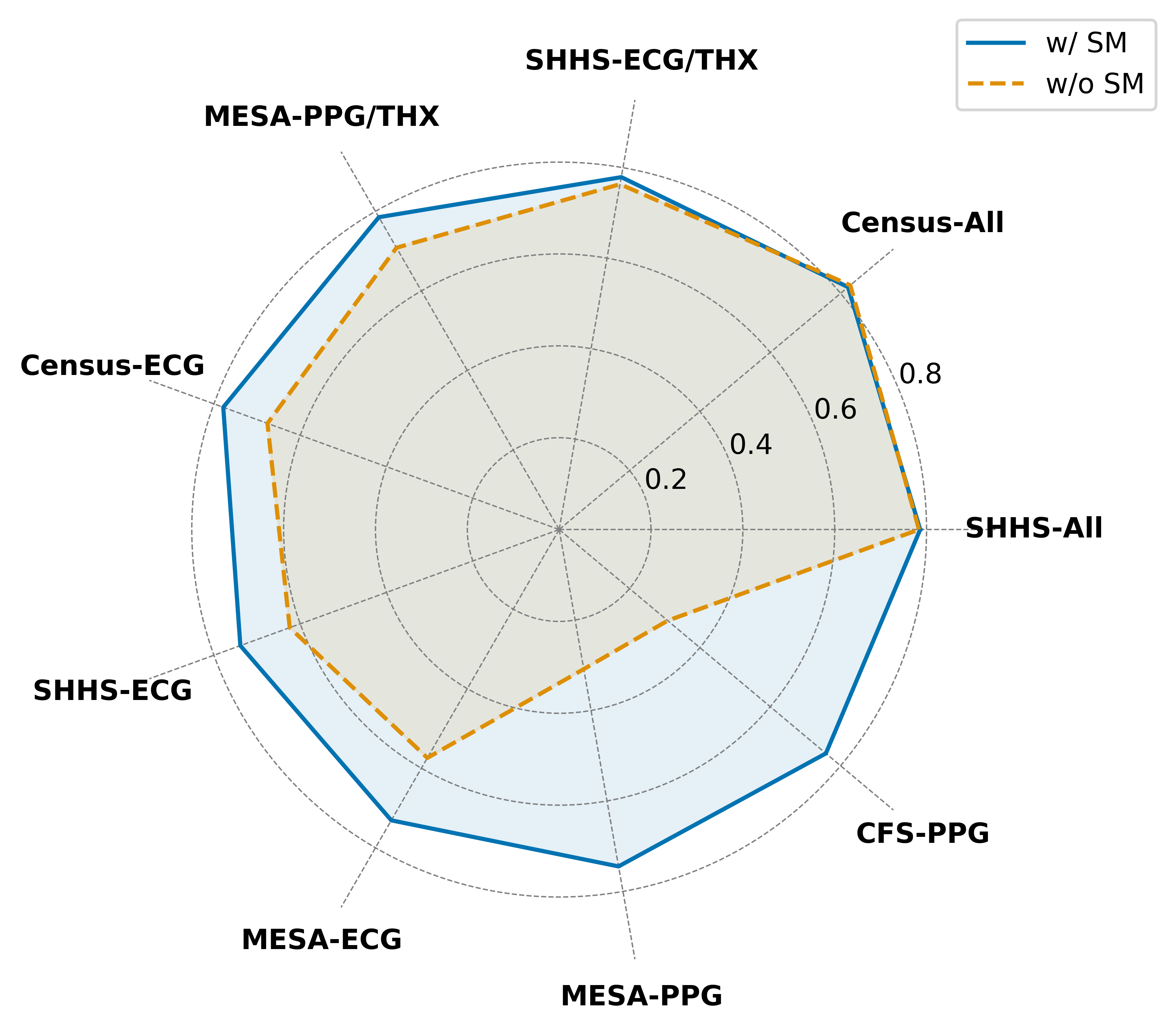}}
\end{figure}

\subsection{Comparison with prior work}
In \Cref{table:wav2sleep:priorcomparison}, we compare the performance of wav2sleep after training on all modalities with prior models trained on specific modalities. We follow the exclusion criteria of \cite{jones_expert-level_2024} and compare against prior work that has explicitly reported the use of distinct training, validation and test sets. Across multiple datasets and combinations of test-time modalities, the wav2sleep model outperforms existing methods for sleep staging from cardio-respiratory signals.
\setlength{\tabcolsep}{0.4em} 
\begin{table}[htb]
\floatconts
{table:wav2sleep:priorcomparison}
{\caption{Comparison of cardio-respiratory sleep staging methods for different test-time modalities $\mathcal{S}_{\,\text{Test}}$.}}
{\tabspace\footnotesize
\begin{tabular}{@{}lllll@{}}
\toprule
Dataset & $\mathcal{S}_{\,\text{Test}}$ & Method  & $\kappa_{T}$ & $\text{Ac}_{T}$\\ \midrule
SHHS & ECG,\,THX & \cite{bakker_estimating_2021}$^\dag$ & 0.64 & 76.7 \\
&  & \cite{carter_sleepvst_2024}  & 0.75 & 83.0   \\ 
& & wav2sleep  & \textbf{0.78} & \textbf{85.0}  \\
\cmidrule{2-5}
&  ECG & \cite{sridhar_deep_2020} & 0.66 & 77.0\\
&  & wav2sleep & \textbf{0.74} & \textbf{82.3}\\
\midrule[0.5pt]
MESA & PPG,\,THX & \cite{bakker_estimating_2021}$^\dag$  & 0.68 & 79.8 \\
   &  & wav2sleep & \textbf{0.78} & \textbf{86.2}\\ 
\cmidrule{2-5}
   & ECG,\,THX & \cite{carter_sleepvst_2024} & 0.77  & 85.2 \\ 
    &   & wav2sleep & \textbf{0.78} & \textbf{86.1} \\ 
\cmidrule{2-5}
  & ECG & \cite{sridhar_deep_2020} & 0.69 & 80.0\\
  &  & wav2sleep & \textbf{0.73} & \textbf{82.8}\\
\midrule[0.5pt]
  Census & ECG & \cite{jones_expert-level_2024}$^\ddag$ & 0.77 & - \\
      &  & wav2sleep  & \textbf{0.78} & \textbf{84.8}\\
\bottomrule
\multicolumn{5}{l}{\scriptsize{Additional model inputs: $^\dag$Nasal airflow,$^\ddag$age and sex.}}
\end{tabular}
}
\end{table}

\section{Conclusions}
\label{section:wav2sleep:conclusions}
In this paper, we have introduced wav2sleep, a deep learning model for automated sleep stage classification that can operate on a variable number of input modalities during training and inference. After joint training on over 10,000 nights of publicly available data from six heterogeneous datasets, this single, unified model leads to improved performance compared to direct training and transfer learning methods across a range of test-time modalities and datasets. Our work further improves the accuracy of sleep staging across a range of important modalities, such as ECG, PPG and respiratory signals, bringing accurate, low-cost sleep monitoring from less obtrusive contact sensors closer to clinical practice. 

\paragraph{Future Work} We have focused on learning from cardio-respiratory signals since sleep staging from these modalities is of particular interest. However, using additional signals such as the EEG may help to further improve the quality of the learnt representations. Finally, the generalised architecture of wav2sleep, particularly the ability to jointly train it on heterogeneous, multi-modal time-series, means that it could be used to complement unsupervised approaches, e.g.~\citep{thapa_sleepfm_2024}.

\acks{This work was supported by the EPSRC Centre for Doctoral Training in Autonomous Intelligent Machines and Systems [EP/S024050/1]. The research was carried out at the National Institute for Health and Care Research (NIHR) Oxford Biomedical Research Centre (BRC). The authors would like to acknowledge the use of the University of Oxford Advanced Research Computing (ARC) facility in carrying out this work. Figure 1 was created with BioRender.com. We kindly thank the National Sleep Research Resource for granting access to the datasets used.}

\appendix

\section{Additional results and discussion}\label{section:wav2sleep:appendix:results}
\subsection{Stochastic masking trade-offs}\label{section:wav2sleep:appendix:results:tradeoff}
As shown in \Cref{table:wav2sleep:ecg}, we observed a small trade-off between ECG and PPG performance using our stochastic masking approach. During training, there are a finite number of optimisation steps before the model begins to overfit. This results in a small performance trade-off between different test-time modality combinations depending on the masking parameters, which determine the relative frequency of modalities observed during training. For example, by increasing the PPG masking probability $p(m_{\text{PPG}})$ to 0.2 we found that the kappa values slightly decreased by 0.01-0.02 across datasets when using only the PPG at test-time, but increased by around the same amount using just the ECG.

A similar effect was noted in the similar approach of Hierarchical Channel Sampling~\citep{bao_channel_2024}, where performance was best for combinations of channels that were most frequently sampled during training. Our stochastic masking procedure means that ECG-only examples are infrequently sampled during training, accounting for less than 1 example per batch on average for datasets that have all four signals available.  Improvements to the stochastic masking procedure, stronger regularisation, and/or a larger batch size (see \Cref{section:appendix:wav2sleep:instance}) may help to address this in future work.

\subsection{Signal noise}\label{section:wav2sleep:appendix:results:noise}
\Cref{fig:wav2sleep:ecg_quality_plot} shows the performance of wav2sleep on the MESA test set for different test-time modalities, grouped by ECG signal quality.\footnote{As measured by the `quecg5' metadata variable.} Here we can observe how the use of multiple input modalities provides improved redundancy. When the ECG signal is of poor quality, the use of additional signals, e.g. THX, helps to maintain good performance.
\begin{figure}[htb]
\floatconts
  {fig:wav2sleep:ecg_quality_plot}
  {\caption{Performance of wav2sleep on the MESA test set, grouped by ECG signal quality index.}}
{\includegraphics[width=1.0\linewidth]{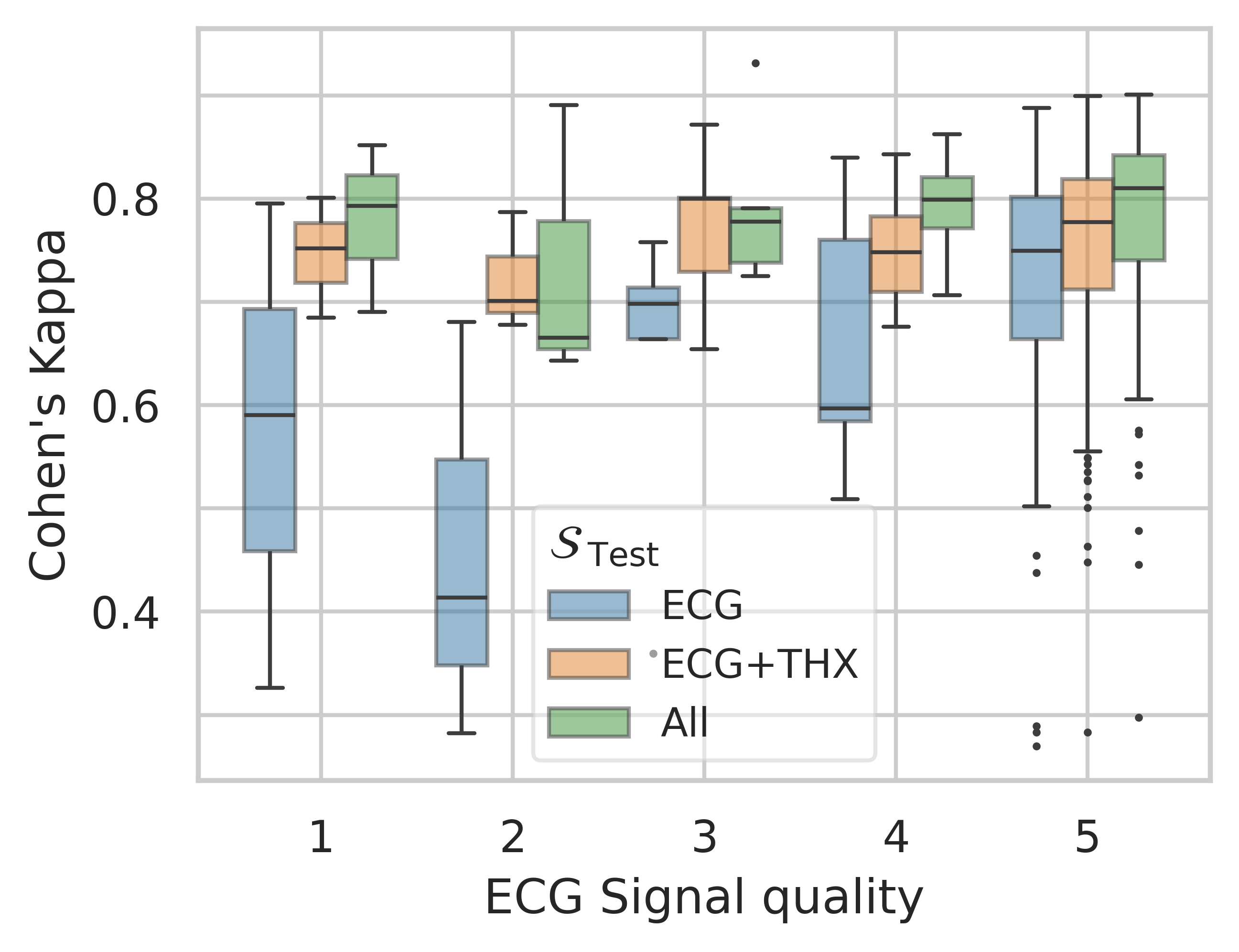}}
\end{figure}

\subsection{Varying test-time modalities}\label{section:wav2sleep:appendix:results:modalities}
After joint training on all datasets and input modalities ($\mathcal{S}_{\,\text{Train}}=\text{All}$), the performance of the wav2sleep model for different test-time modalities $\mathcal{S}_{\,\text{Test}}$ is listed in \Cref{table:wav2sleep:all}.
\begin{table*}[htb]
\floatconts
{table:wav2sleep:all}
{\caption{wav2sleep performance ($\kappa_{T}$) for different test-time modalities.}}
{\tabspace
\small
\begin{tabular}{@{}lcccccccc@{}}
\toprule
 & \multicolumn{8}{c}{Dataset} \\ 
\cmidrule(l){2-9}
$\mathcal{S}_{\,\text{Test}}$ & SHHS & MESA & WSC & CHAT & CFS & CCSHS & MROS & Census\\ 
\midrule
All$^\dag$ & 0.786 & 0.796 & 0.737 & 0.836 & 0.803 & 0.857 & 0.805 & 0.821\\
ECG,THX & 0.779 & 0.783 & 0.728 & 0.827 & 0.802 & 0.854 & 0.796 & 0.812\\
ECG & 0.739 & 0.731 & 0.689 & 0.800 & 0.784 & 0.833 & 0.750 & 0.783\\
PPG & - & 0.742 & - & 0.793 & 0.763 & 0.832 & - & -\\
\bottomrule
\multicolumn{7}{@{}l@{}}{\scriptsize $^\dag$ABD+THX+ECG (+PPG for MESA, CHAT, CFS and CCSHS)} \\
\end{tabular}
}
\end{table*}

\subsection{Example hypnograms}\label{section:wav2sleep:appendix:hypnograms}
\Cref{fig:wav2sleep:arrhythmia_hypnograms} shows example sleep hypnograms generated by the wav2sleep model using different test-time modalities. This night of data\footnote{Session ID: wsc-visit2-12529-nsrr} corresponds to an elderly male with diagnosed cardiac arrhythmia and mild sleep apnea. Visually, the ECG is of good quality, however, using the ECG as the sole input to the model results in poor agreement with expert-annotated sleep stages. The addition of the thoracic signal results in a significant performance improvement.

Notably, we found that when using the ECG as the sole input, performance \textbf{improves} with apnea severity for subjects with cardiac arrhythmia (see \Cref{fig:wav2sleep:arrhythmia_apnea_plots}). This is in contrast to the general trend seen in prior work that performance tends to decrease with apnea severity~\citep{korkalainen_accurate_2020}. We hypothesise that, for subjects with arrhythmia, the model may mistake heart rate variability (HRV) caused by arrhythmia for HRV caused by the more common condition of sleep apnea~\citep{penzel_is_2003}. In turn, this may confound the learnt mapping between physiological features and sleep stages i.e. the mapping $g$ described in \Cref{section:wav2sleep:motivation}.
\begin{figure*}[htb]
\floatconts
  {fig:wav2sleep:arrhythmia_hypnograms}
  {\caption{\textbf{Example sleep hypnograms for a subject with diagnosed cardiac arrhythmia.} (top) Annotated by a human expert using the PSG recording. (middle) Produced by the wav2sleep model using ECG and THX signals ($\kappa_{T} = 0.74$). (bottom) Produced by the wav2sleep model using the ECG signal ($\kappa_{T}=0.19$).}}
{\includegraphics[width=1.0\linewidth]{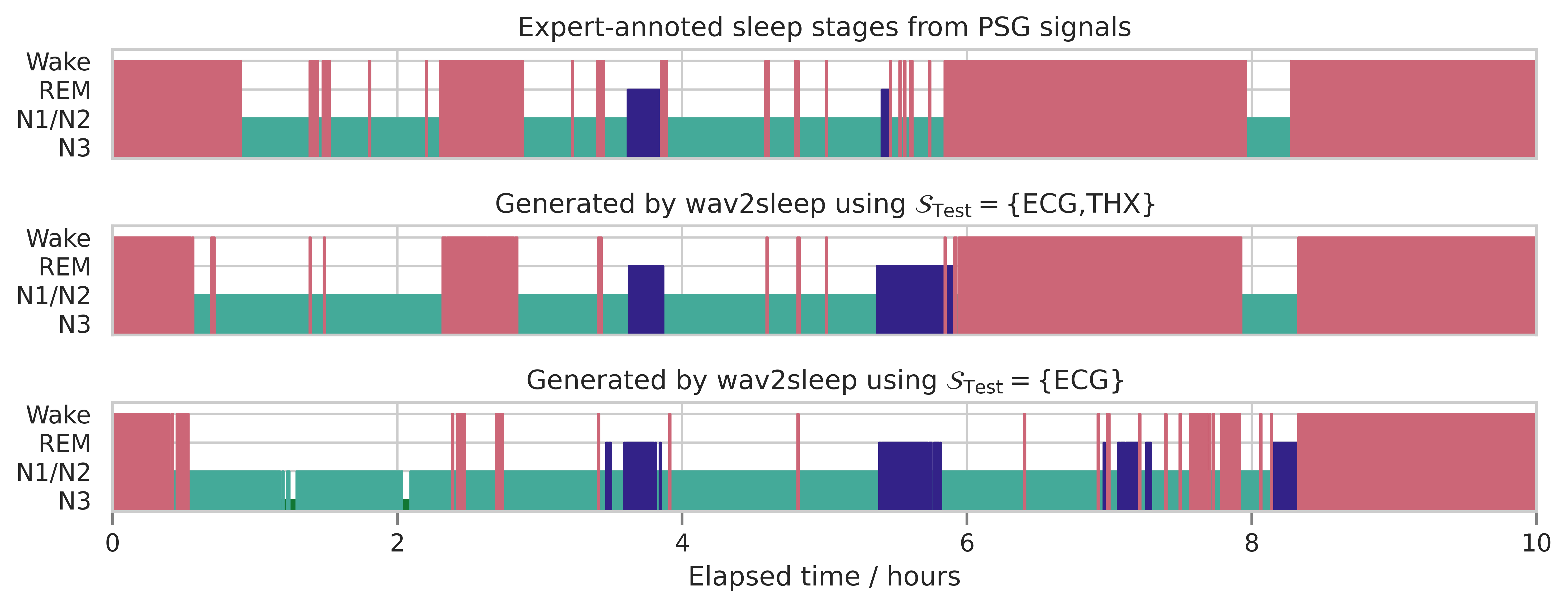}}
\end{figure*}

\begin{figure*}[htb]
\floatconts
  {fig:wav2sleep:arrhythmia_apnea_plots}
  {\caption{\textbf{Performance of wav2sleep on the WSC test set, grouped by apnea severity and arrhythmia}. For subjects with arrhythmia, and using only the ECG at test-time, performance \emph{improves} with apnea severity. This is in contrast to the trend that performance decreases with apnea severity seen using other modalities and for subjects without diagnosed arrhythmia.}}
{\includegraphics[width=0.9\linewidth]{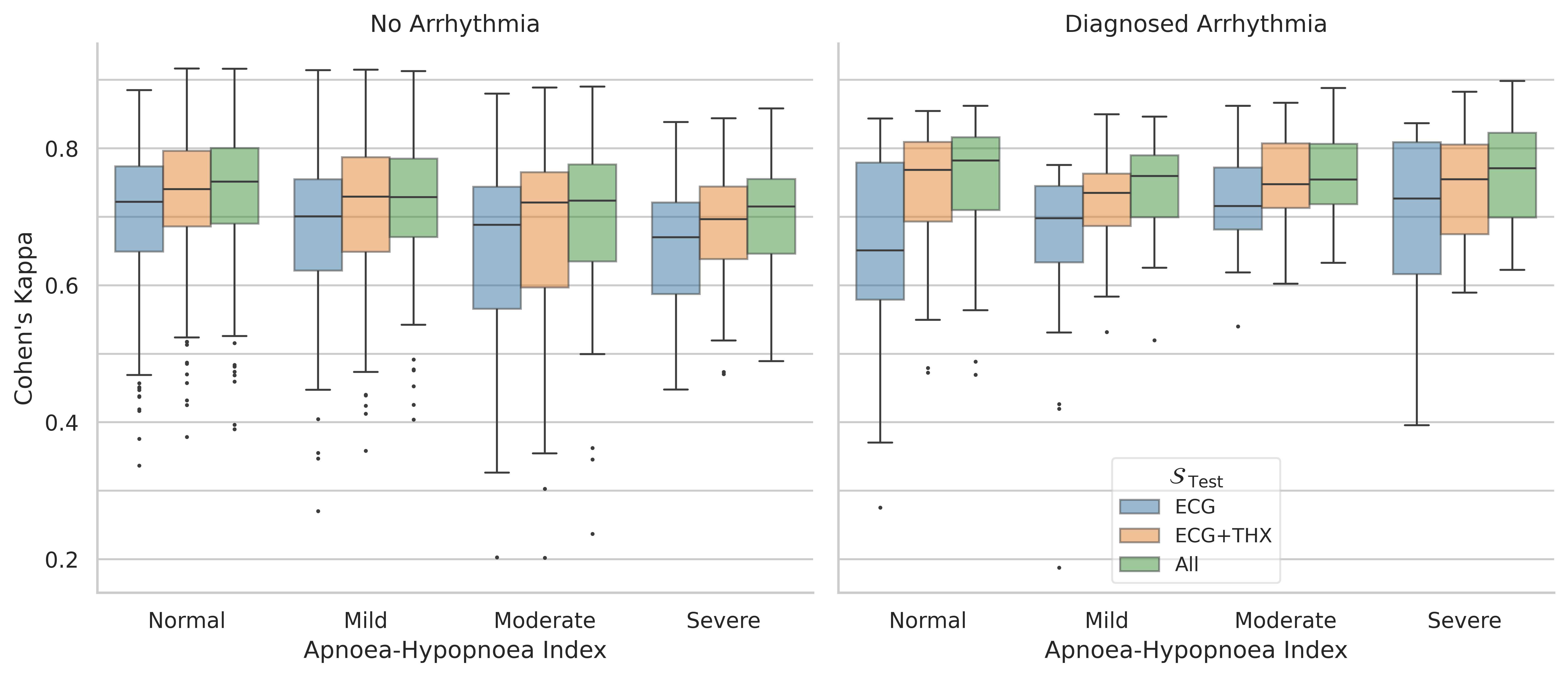}}
\end{figure*}

\section{Dataset processing}\label{section:wav2sleep:appendix:dataset_processing}
\subsection{Scoring exclusions}
Because of signal quality issues, some of the recordings in each dataset only have binary sleep--wake annotations, rather than full AASM (Wake, N1, N2, N3, REM) sleep stages. For these recordings, \emph{all} sleep stages are typically assigned to the same integer as `N2' sleep. This means that these labels should not be used for training or evaluation of multi-class sleep staging models. Where available, we used the harmonised `nsrr\_flag\_spsw' metadata variable produced by the National Sleep Research Resource to exclude these recordings. Otherwise, we checked for the existence of either N1, N3 or REM sleep labels.

\subsection{Construction of test sets}
From the CFS and CHAT datasets, we created our validation and test sets using recordings where the PPG signal was available, to enable evaluation across all combinations of modalities. The remaining recordings (with and without PPG available) were used for training. We used recordings from the non-randomised (single night per participant) arm of the CHAT dataset for our test set. Similarly to \cite{carter_sleepvst_2024}, we selected 1000 nights for our SHHS test set by randomly choosing 500 participants who participated in both visits. For the WSC dataset, we selected 2 recordings from 250 participants who had undertaken at least 2 visits to form our test set of 500 recordings; additional recordings from participants in the test set were excluded. Our decision to use test sets with two recordings per person for the SHHS and WSC datasets was taken to maximise data usage while avoiding the same participant appearing simultaneously in both the training and test sets. In future work, these sets could be used for additional analysis such as the variation in performance with age after controlling for identity. To enable evaluation on the census-balanced test set proposed by \cite{jones_expert-level_2024}--which uses recordings from CCSHS, CFS, CHAT, MESA and WSC--we excluded their test set recordings from our training and validation sets.

\section{Model design}\label{section:appendix:wav2sleep:model_design}
Here we describe additional experiments and observations that informed the design and hyper-parameters of the wav2sleep model.  Hyper-parameter search was informed by the minimum validation loss $\mathcal{L}_{\text{Min}}$ during initial experiments.

\subsection{Signal Encoders}\label{section:appendix:wav2sleep:instance}
We found that using instance normalisation~\citep{ulyanov_instance_2017} within the signal encoders and layer normalisation~\citep{ba_layer_2016} in the sequence mixer improved training stability and performance. Because of our stochastic masking procedure, the number of examples of a signal within a batch will often be much smaller than the actual batch size, increasing the variance of statistics used by the more common approach of batch normalisation~\citep{ioffe_batch_2015}.

\subsection{Epoch Mixer}
We evaluated two designs for the epoch mixer:
\begin{enumerate}
    \item A small transformer encoder (TE) i.e. our best-performing approach.
    \item A linear concatenation and projection layer, handling variation in the available inputs with zero-padding.
\end{enumerate}
The attention-based epoch mixer achieved a lower validation loss and higher Cohen's $\kappa$ values across multiple datasets and modalities.
\begin{table}[htb]
\floatconts
{table:wav2sleep:epoch_mixer_design}
{\caption{Performance comparison of epoch mixer designs for sample validation set metrics.}}
{\tabspace
    \small
    \begin{tabular}{@{}lcccc@{}}
    \toprule
    & & \multicolumn{3}{c}{Dataset-modality $\kappa_{T}$}\\
    Design & $\mathcal{L}_{\text{Min}}$ & \footnotesize{C-ECG} & \footnotesize{M-PPG} & \footnotesize{S-ECG}\\ \midrule
    Linear & 0.351 & 0.770 & 0.741 & 0.719\\
    TE & 0.349 & 0.773 & 0.743 & 0.723\\ 
    \bottomrule
    \multicolumn{5}{l}{\scriptsize{C - Census. M - MESA. S - SHHS}}
    \end{tabular}
}
\end{table}

\subsection{Sequence Mixer}
We evaluated two designs for the sequence mixer:
\begin{enumerate}
    \item A dilated convolutional (DCNN) design, as originally proposed by \cite{sridhar_deep_2020}.
    \item A transformer encoder (TE) with sliding window attention~\citep{beltagy_longformer_2020} and rotary positional embeddings~\citep{su_roformer_2024}.
\end{enumerate}

\begin{table}[htb]
\floatconts
{table:wav2sleep:sequence_mixer_design}
{\caption{Performance comparison of sequence mixer designs for sample validation set metrics.}}
{\tabspace
    \small
    \begin{tabular}{@{}lcccc@{}}
    \toprule
    & & \multicolumn{3}{c}{Dataset-modality}\\
    Design & $\mathcal{L}_{\text{Min}}$ & \footnotesize{C-ECG} & \footnotesize{M-PPG} & \footnotesize{S-ECG}\\ \midrule
    TE & 0.355 & 0.764 & 0.724 & 0.706\\
    DCNN & 0.349 & 0.773 & 0.743 & 0.723\\ 
    \bottomrule
    \multicolumn{5}{l}{\scriptsize{C - Census. M - MESA. S - SHHS}}
    \end{tabular}
}
\end{table}

We found that the DCNN design consistently achieved better performance across different modalities, and converged after fewer training epochs. The hyper-parameters of the dilated convolutional design have been carefully tuned through extensive hyper-parameter search in prior work~\citep{sridhar_deep_2020, kotzen_sleepppg-net_2023}. Though we did perform a basic search over transformer hyper-parameters, such as the number of encoder layers and the context length, performing extensive tuning was deemed unnecessary given the results achieved using a convolutional design, and outside the scope of this paper. Using a well-tuned mixture of local and global attention~\citep{beltagy_longformer_2020} may yet lead to superior performance using a transformer-based architecture, but is left for future work.

Finally, it is worth noting that our implementations of both stochastic masking and local attention relied on naive masking of the attention matrix using PyTorch~\citep{paszke_pytorch_2019}. Using optimised sparse kernels (e.g. from xformers~\citep{xFormers2022}) could provide significant speed-ups and efficiency gains on modern GPU architectures, making a transformer-based architecture a more attractive option for training on an even larger quantity of data.

\section{Model training}\label{section:wav2sleep:appendix:training}
Model parameters $\theta$ were found by minimising the unweighted cross-entropy loss between one-hot encoded labels $\bm{y}_{1:T}\in\mathbb{R}^{C\times T}$ and output probabilities $\bm{p}_{1:T}\in\mathbb{R}^{C\times T}$. For each night of data, the total cross-entropy loss is given by:
\begin{align}
    &\mathcal{L}_{\theta}(\bm{y}_{1:T}, \bm{p}_{1:T}) = - \sum_{i=1}^{C}\sum_{j=1}^T \left(\bm{y}_{1:T}\odot \log(\bm{p}_{1:T})\right)_{ij}
\end{align}
where $\odot$ denotes the Hadamard product.

\paragraph{GPU training}
Experiments were performed using a computing cluster containing multiple GPU architectures. Gradient accumulation was used to ensure a consistent effective batch size of 16, using the largest batch size that could fit on the particular GPU(s) used in a given experiment. Using a single NVIDIA A100, the actual batch size was 4 samples, and each epoch took 21 minutes, resulting in an average training time of around 10 hours.

\end{document}